\documentclass[lettersize,journal]{IEEEtran}
\usepackage{amsmath,amsfonts}

\usepackage{algorithmic}
\usepackage{algorithm}
\usepackage{array}
\usepackage[caption=false,font=normalsize,labelfont=sf,textfont=sf]{subfig}
\usepackage{textcomp}
\usepackage{stfloats}
\usepackage{url}
\usepackage{verbatim}
\usepackage{graphicx}
\usepackage{cite}
\usepackage[font=small,skip=2pt]{caption} 

\begin{document}
\title{

A Hermetic, Transparent Soft Growing Vine Robot System \\for Pipe Inspection

}

\author{
William E. Heap$^{1}$, Yimeng Qin$^{1}$, Kai Hammond$^{1}$, Anish Bayya$^{2}$, Haonon Kong$^{1}$, and Allison M. Okamura$^{1}$

\thanks{This research was supported in part by National Science Foundation Award 2345769, and the Stanford University Doerr School of Sustainability. (\it{Corresponding author: William Heap, wheap@stanford.edu})}

\thanks{$^{1}$W. E. Heap, Y. Qin, K. Hammond, H. Kong, and A. M. Okamura are with the Department of Mechanical Engineering, Stanford University, 94305, California, United States of America}

\thanks{$^{2}$A. Bayya is with the Department of Aeronautics and Astronautics, Stanford University, 94305, California, United States of America}}%



\maketitle

\begin{abstract}
Rehabilitation of aging pipes requires accurate condition assessment and mapping far into the pipe interiors. Soft growing vine robot systems are particularly promising for navigating confined, sinuous paths such as in pipes, but are currently limited by complex subsystems and a lack of validation in real-world industrial settings. In this paper, we introduce the concept and implementation of a hermetic and transparent vine robot system for visual condition assessment and mapping within non-branching pipes. This design encloses all mechanical and electrical components within the vine robot's soft, airtight, and transparent body, protecting them from environmental interference while enabling visual sensing. Because this approach requires an enclosed mechanism for transporting sensors, we developed, modeled, and tested a passively adapting enclosed tip mount. Finally, we validated the hermetic and transparent vine robot system concept through a real-world condition assessment and mapping task in a wastewater pipe. This work advances the use of soft-growing vine robots in pipe inspection by developing and demonstrating a robust, streamlined, field-validated system suitable for continued development and deployment.


\end{abstract}

\begin{IEEEkeywords}
Soft robot applications, soft robot materials and design
\end{IEEEkeywords}

\section{Introduction}
\IEEEPARstart{L}{arge} portions of the millions of kilometers of pipes that span the globe are in disrepair, posing safety risks, continuously leaking resources, and contaminating the environment \cite{pipe_risks, water_report}. Rehabilitation of these pipes relies on condition assessment and mapping data \cite{pipe_rehabilitation_decisions, pipe_risk_rehab_2}. Condition assessment involves collecting qualitative or quantitative data on the physical state of the pipe through methods such as visual inspections, thickness measurements, and leak detection. Mapping involves the generation of geo-spatial location data for the pipe itself, as well as the locations of points of interest such as blockages, cracks, or collapsed sections. While accurate maps are commonly available for modern pipes, many older pipes installed decades or even centuries ago do not have accurate records of their locations \cite{bad_pipe_maps}. Additionally, maps created during the pipe installation cannot provide the location of additional features added over time. 

\begin{figure}[t]
\centerline{\includegraphics[width=85mm]{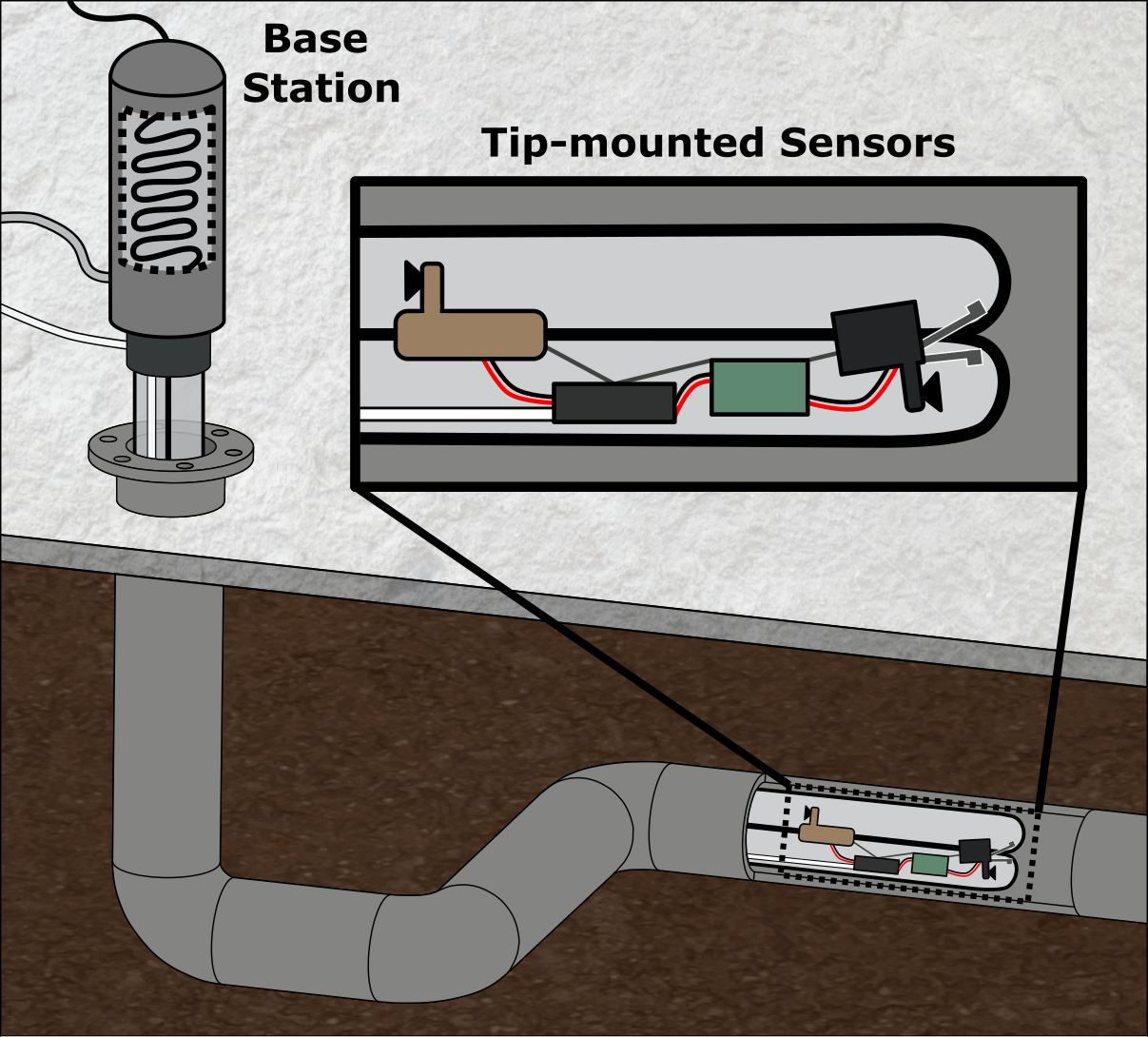}}
\caption{A visualization of the hermetic, transparent soft growing vine robot system being used to inspect a buried pipe. The vine robot enters the pipe from the top left, and the inset image shows the electronics and mechanical equipment enclosed and hermetically sealed at the tip of the robot.}
\label{fig:firstfig}
\vspace{-0.5cm}
\end{figure}

Condition assessment and mapping can be performed from the exterior or interior of a pipe. Exterior methods eliminate the need to insert equipment into the pipe, but buried pipes are difficult to access. \cite{external_pipe_scan, external_pipe_methods}. 
The alternative is to send tools or robots into the interior of the pipe. These options can be grouped based on propulsion method. The first group, consisting of scew/helical robots, wheeled/tracked robots, legged robots, snake-like robots, and worm-like robots, all rely on frictional contact with the pipe walls to generate propulsive forces \cite{pipe_inspection_review, in_pipe_review}. Rigid versions of these robots must have high degrees of freedom and multiple actuators to handle changing pipe diameters and sharp turns. Meanwhile, soft robotic versions can reduce system complexity by employing as few as a single actuator \cite{single_actuator_soft}. However, they also typically rely on friction, limiting their ability to propel long tethers or sensor payloads in slick or vertical pipe sections. 

Alternatively, drones \cite{drone}, flow-propelled systems \cite{smart_ball, pigs_review}, underwater vehicles \cite{fish_robot, underwater_vehicles}, borescopes \cite{borescope}, and soft growing vine robots \cite{scirobo_vine} do not rely on friction for propulsion force. However, drones are limited by their size and maximum propulsive force, while flow-propelled systems and underwater vehicles 
can be mechanically complex and rely on the pipe being filled with fluid. Borescopes are mechanically simple, but cannot navigate sinuous pipes because increasing the propulsion force causes the borescope body to buckle. 

Soft growing robots can generate large propulsion forces by pressurizing a tube with inverted material, causing the robot to lengthen from its tip \cite{small_vine, largevine}. The propulsion forces are not affected by environmental conditions and require only an air compressor and an airtight tube, making vine robots a promising method for in-pipe inspection. 
Prior work on vine robot systems has addressed  navigation of pipe junctions, mounting sensors on the outside of the vine robot tip or body, and controlling length. These solutions have introduced  additional actuators, complex mechanisms, and heavy subsystems \cite{vine_inpipe, roboa, largevine,internal_reeling_steering}. 
Externally mounted sensors also require waterproofing, protection from fouling, and additional electronics for wireless data transfer. Alternate methods that run a wired camera or other tools through the center of the vine robot are less complex, but still require sensors to be exposed to the environment, can limit the maximum deployable length of the robot, and have not yet been demonstrated in real-world pipe-inspection applications \cite{scrunching_vine, origami_vine}.  

While useful in some situations, the advanced vine robot capabilities described above are unnecessary for many pipe inspection and mapping tasks. Thus, we present a hermetic and transparent soft growing vine robot system that offers a promising trade-off between generalizability and system complexity. This is possible for two reasons. First, vine robots that have all components enclosed within their airtight body can still be effective for common but challenging pipe condition assessment and mapping tasks, as the use of a transparent body material enables visual sensing of the environment from inside the vine robot body.
Second, by enclosing all components within the vine robot body, we enable a series of system simplifications, resulting in a more robust and practical system for real-world pipe inspection tasks.
In this paper, we present the design, implementation, and testing of a hermetic and transparent vine robot system optimized for assessment and mapping tasks within sinuous pipes without junctions. We also investigate methods of passively transporting sensors with a fully enclosed mechanism, a previously unexplored area of research. The contributions of this work are: (1) the design and implementation of a hermetic and transparent vine robot system; (2) the modeling and characterization of a method to efficiently keep sensors at the tip of the vine robot body; and (3) the deployment of the system in a real-world wastewater pipe condition assessment and mapping task.

\section{Design Concept}

Our aim was to develop an efficient solution for condition assessment and mapping tasks within long, sinuous pipes without junctions.  
Here we discuss the rationale for a hermetic and transparent vine robot system, which differs from prior vine robot systems in three interconnected ways: 

\emph{1. All mechanical and electronics equipment is housed within a transparent vine robot body.} While prior vine robot bodies have been fabricated from transparent materials, the transparency has not been used to enable internally mounted cameras. Fully enclosing all equipment provides three advantages. First, no wireless communication equipment is needed to transfer data across the physical barrier of the vine robot body. Second, all mechanical and electronic components are protected by the airtight, soft vine robot body. This eliminates the need for additional waterproofing or lens-clearing features, minimizes the risk of rigid elements snagging or breaking against pipe features, and allows the vine robot to pass through apertures smaller than its fully inflated diameter. Lastly, transparent thin-film plastic tubes can be commercially purchased in long lengths and varying configurations due to their frequent use in packaging products. 

\emph{2. A passively adapting, fully enclosed tip mount transports sensing and communication equipment.} Vine robots navigate an environment via tip extension rather than locomotion, so sensors cannot be permanently attached to a single location on the vine robot body. Because the tail of a vine robot travels at twice the speed of its tip, tip mounts that ensure fixation at the changing tip of a vine robot while remaining sealed inside require that the tail slide through a clamping mechanism. However, our modeling and experimental demonstrations in Sections \ref{section:modeling} and \ref{section:experimentaltesting} show that much propulsion force is lost in continuously sliding a tip-mount along the tail. More efficient tip mounts designs could introduce actuators and parts exposed to the environment. Instead, we developed a passively adapting enclosed tip mount whose clamping force automatically adjusts based on its proximity to the tip of the vine robot, minimizing loss of propulsive force. 

\emph{3. The vine robot is deployed from a canister-style base station that has no moving parts.} The canister-style base station stores the vine robot tail and tether material within a pressurized chamber, similar to conventional base stations. Tethers eliminate the need for batteries and provide operators with live data in wireless signal-denied environments. In our system, vine robot growth and removal are performed manually. To control growth, the operator regulates the vine robot's internal pressure or the amount of tether paid out. To remove the vine robot from a pipe, it is deflated and manually pulled out by its body and tail simultaneously. This base station design eliminates the need for the powerful actuators and heavy structures required for motorized vine robot length control \cite{roboa, largevine, high_pressure_vine}.

\section{System Design and Fabrication}
\label{sec:design}

\begin{figure}[h!]
\centerline{\includegraphics[width=85mm]{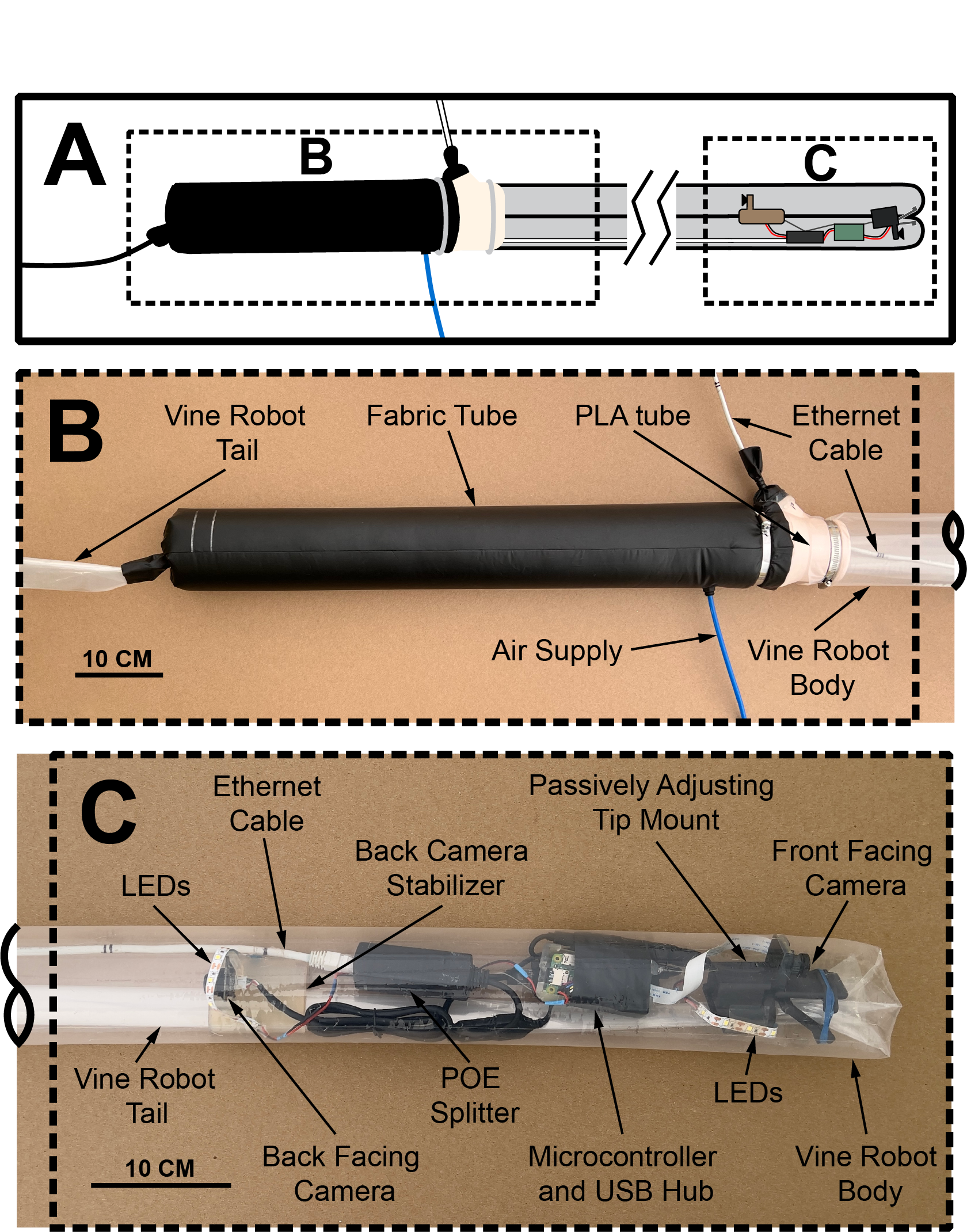}}
\caption{Enclosed vine robot system. (A) Schematic of the entire system; the middle of the vine robot body is hidden. The relative locations of the subsystems B and C are also shown. (B) The canister-style base station and key features. (C) The distal end of the enclosed vine robot, with the internal electrical and mechanical elements housed within the vine robot body.}
\label{fig:robotdiagram}
\vspace{-0.5cm}
\end{figure}

Our implementation, shown in Fig. \ref{fig:robotdiagram}, is be subdivided into four subsystems: (1) the transparent vine robot body, (2) canister-style base station, (3) passively adapting enclosed tip mount, and (4) electronics, communications, and mapping. We eliminated the need for active steering elements, using vine robots' ability to passively navigate non-branching pipes with bends via environmental interaction \cite{GreerIJRR2020}.

\subsection{Transparent Vine Robot Body}
\label{section:designvinerobotbody}
The vine robot body (Figs. \ref{fig:robotdiagram}B-C) is formed by an airtight, 0.18 mm thick, 88 mm diameter low-density polyethylene (LDPE) tube (Uline). The thickness was chosen to maximize puncture and abrasion resistance while maintaining optical clarity. The vine body houses the mechanical and electronics equipment, and is coupled to the base station to form an airtight chamber. To reduce friction and material deformation as the vine robot tail passes through the base station and tip mount, we fold the tube in a serpentine pattern before inserting it into the base station. 

\subsection{Canister-Style Base Station}

As the vine robot grows, the vine robot tail and tether must be continuously fed through the pressurized vine robot body. Our re-usable base station consists of a thermoplastic polyurethane-coated 70 denier ripstop-nylon fabric (Quest Outfitters) sealed into a tube 49 mm in diameter and 0.9 m long using an ultrasonic welder (Vetron 5064). As seen in Fig. \ref{fig:robotdiagram}B, one end of the fabric tube and the open end of the vine robot body are attached with hose clamps to each end of a 3D printed PLA tube. The other end of the fabric tube is cinched closed with cable ties, forming an airtight chamber. The PLA tube also has two ports, one for connecting an air supply tube, and one for the tether (an Ethernet cable) to be fed through. A short fabric sleeve is adhered to the Ethernet cable port, allowing the port to be manually closed or cinched with cable ties to maintain pressure within the vine robot. 

Up to 50 m of vine robot tail material can fit within the fabric tube. The free end of the vine robot tail is passed through the cinched end of the fabric tube (Fig. \ref{fig:robotdiagram}B), allowing additional tail material to be inserted as needed. During operation, Ethernet cable is fed into the vine robot body. Due to the optically clear vine robot body, the operator can observe when all the Ethernet cable slack is taken up, and feed more through the base station. 

\subsection{Passively Adapting Enclosed Tip Mount}

\begin{figure}[t]
\centerline{\includegraphics[width=0.8\columnwidth]{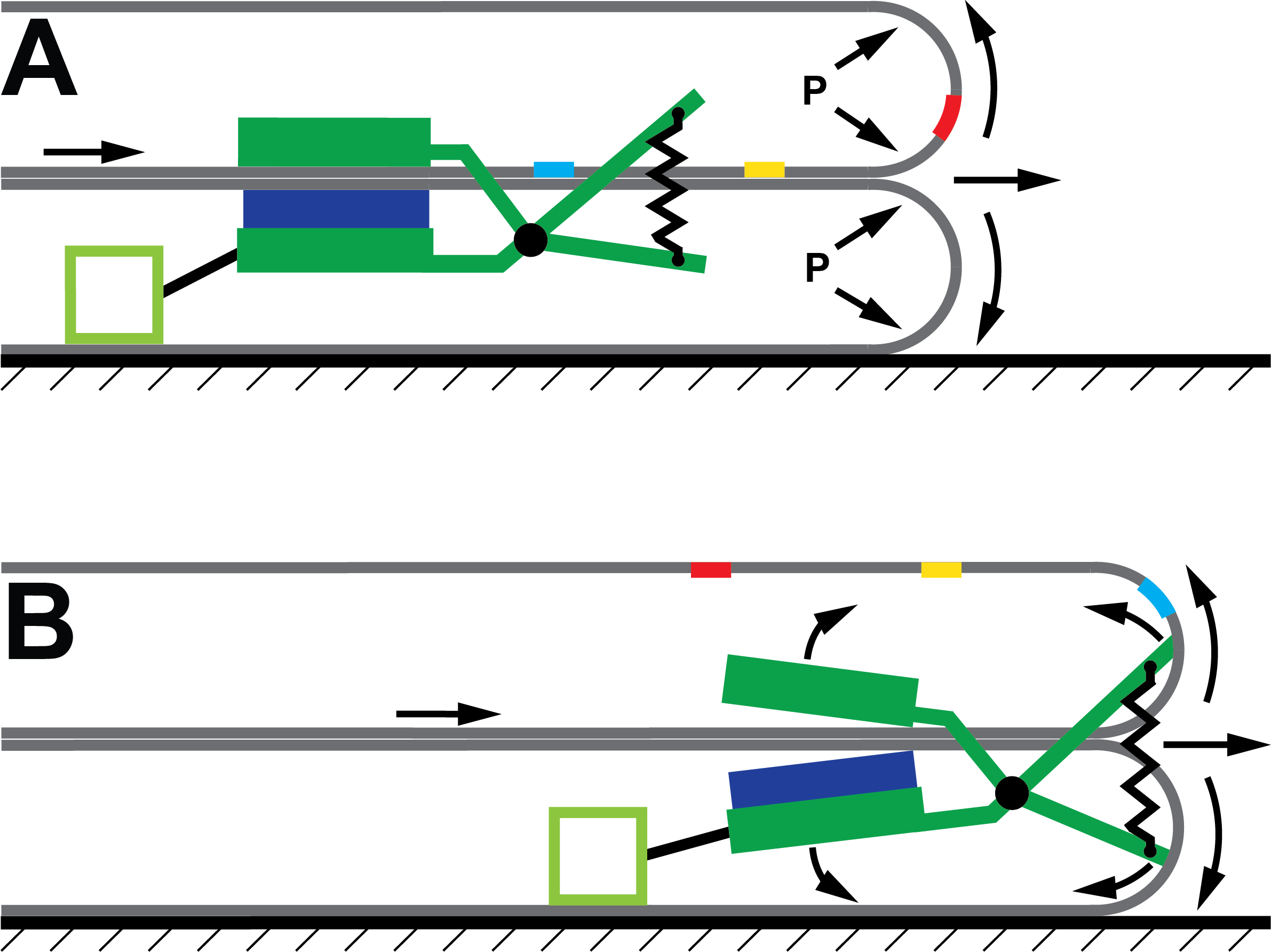}}
\caption{Working principle of the passively adapting enclosed tip mount. (A) The tip mount is offset from the tip of the vine robot, and clamps to the vine robot tail. As the vine robot grows to the right, tail material and attached tip mount are pulled to the tip. (B) When the tip mount reaches the tip of the vine robot, its clamp is forced open to allow the vine robot's tail material to slide through.}
\label{fig:tipmountgrowth}
\vspace{-0.5cm}
\end{figure}

The tip mount is a 3D printed part (Onyx, Markforged) that uses a scissor-like mechanism with elastic bands to clamp onto the vine robot tail and keep the electronics equipment and tether at the tip of the vine robot body. The clamp surface is a high-friction silicone sheet (Dycem). As the vine robot grows and pulls tail material to its tip, the tip mount and attached equipment are pulled along with it (Fig. \ref{fig:tipmountgrowth}A). The tip mount stops once it reaches the tip of the vine robot, and its clamp prevents additional tail material from passing through the tip mount and everting. The vine robot can then only grow when the net propulsive force on the vine robot tail exceeds the friction force from the tip mount's clamp. The curvature of the vine robot tail material as it everts pushes apart the two arms of the tip mount and reduces the required propulsive force to overcome the clamping force (Fig. \ref{fig:tipmountgrowth}B).

\subsection{Electronics, Communications, and Mapping}

The vine robot's enclosed electronics system (Fig. \ref{fig:robotdiagram}C) transmits data, while the operator’s computer receives data. Fig. \ref{fig:robotdiagram}C shows that the internal electronics is composed of two LED-illuminated cameras (Adafruit IMX708 and IMX291 wide-angle fisheye cameras) that provide visual feedback of both forward and backward directions, a microcontroller unit (MCU; Raspberry Pi Zero 3W) with a USB hub serving as the centralized local computing unit, an inertial measurement unit (IMU; Adafruit BNO055) for 9-axis orientation sensing, including acceleration, angular velocity, and magnetic field, and a Power-over-Ethernet (PoE) splitter (UCTRONICS 5V model) that delivers power and communication. The MCU captures video streams from the two cameras (720p at 30 fps via USB and MIPI interfaces) and IMU data via the I²C bus, then transmits aggregated sensor packets to the local computer over an IEEE 802.3 local area network (LAN). On the receiver side, a laptop equipped with a PoE injector (PoE-M911-N 54V 30W model) provides power to the tip-mounted electronics and receives the raw sensor data.

To reconstruct the pipe geometry, we combined the robot’s orientation measurements with manually labeled odometry data, obtained by placing markers along the Ethernet cable, where each marker denoted a key location in the pipe (such as the start and end of straight sections and elbows). Using the measured orientation of the robot in the sensor frame, we first transformed it into the world frame—defined with the negative gravity direction as the $z$-axis and magnetic north and west as the $x$ axis and $y$ axis-by constructing a rotation matrix from the accelerometer and magnetometer vectors. With the known odometry of straight pipe sections, the corresponding orientation data, and the measured arc length of elbows, the full 3D pipe geometry can be reconstructed.


\section{Vine Robot Growth Model}\label{section:modeling}

In this section, we model the pressure required to grow a vine robot, including the effects of payload, tether, and tip mounts (and in particular, our passively adapting tip mount).

Vine robot growth is governed by the force balance \cite{apical_modeling}:
\vspace{-0.1cm}
\begin{equation}
\label{eq:simplegrowth}
    CPA \geq F_{\text{eversion}} + T_{\text{tail}} , 
\end{equation}
\noindent
where $C$ is a geometric factor, $P$ is the internal vine robot pressure, $A$ is the cross-sectional area of the vine robot, and $F_{\text{eversion}}$ is the robot tail material's resistance to eversion. $T_{\text{tail}}$ is the resistive tension applied by the vine robot tail, which arises from friction, interactions with the base station, and gravity resisting the motion of the tail. Growth can only occur when $CPA$, the propulsive force generated by the vine robot, equals or exceeds the sum of the resistive forces.

Inversion, the opposite process of growth, occurs when \cite{retraction}: 
\vspace{-0.1cm}
\begin{equation}
\label{eq:inversion}
    T_{\text{tail}} \geq \frac{1}{2}PA + F_{\text{inversion}},
\end{equation}
\noindent
where $F_{\text{inversion}}$ is the robot tail material's resistance to inversion. Due to symmetry between the growth and inversion processes, $C$ has been assumed to be $\frac{1}{2}$ during growth, which is supported by our experimental findings in Section \ref{section:experimentalnotip}.

\subsection{Growth with a Payload and Tether}

A vine robot can transport a payload and tether during growth by directly attaching them to the vine robot tail. Under these conditions, the new growth model (with the substitution $C = \frac{1}{2}$) becomes:
\vspace{-0.1cm}
\begin{equation}
\label{eq:simpletetherpayload}
    \frac{1}{2}PA \geq F_{\text{eversion}} + T_{\text{tail}} + F_{\text{load}} ,
\end{equation}
\noindent
where $F_{\text{load}}$ represents the sum of the frictional, gravitational, and base station-induced forces resisting the motion of the tether and the payload. This expression represents the lowest pressure to grow required for a vine robot of a given material, cross section, tether, spatial configuration, and payload. 

\subsection{Tip Mounts}

Because the vine robot tail travels at twice the velocity of the vine robot tip, loads cannot be statically attached to the vine robot tail. Instead, an enclosed tip mount can be used to transfer force from the vine robot tail to the tether and payload using friction. Here we model (1) a baseline constant force design and (2) our passively adjusting design to determine their impact on vine robot growth.


\subsubsection{Constant Force Enclosed Tip Mount}

\begin{figure}[h!]
\centerline{\includegraphics[width=85mm]{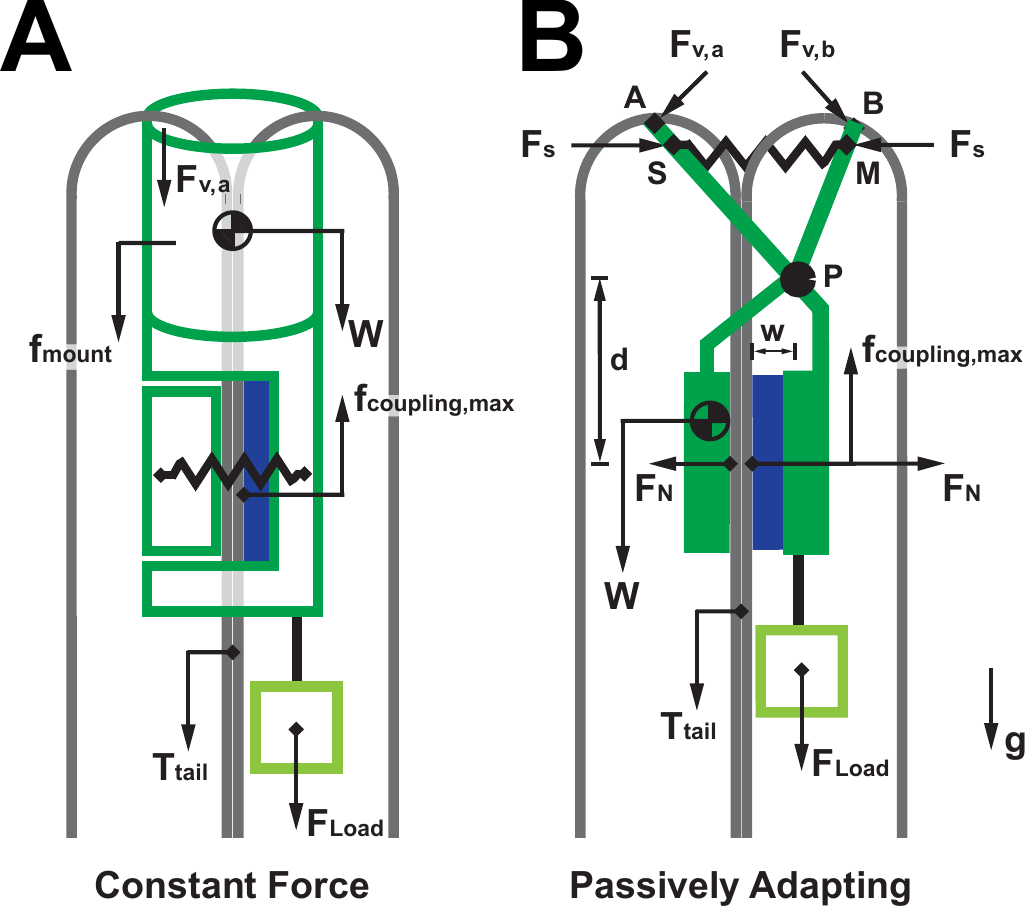}}
\caption{(A) A free-body diagram of a constant force enclosed tip mount design. (B) A free-body diagram of a passively adapting enclosed tip mount design.}
\label{fig:fbds}
\end{figure}

We term the class of enclosed tip mount designs that couple to the tail with a constant friction force ``constant force'' enclosed tip mounts. From the free-body diagram of a constant force enclosed tip mount in Fig. \ref{fig:fbds}A, we see that the tip mount will be pushed away from the vine robot tip so growth can occur when:
\vspace{-0.1cm}
\begin{equation}
    F_{v,a} + F_{\text{load}} + W - f_{\text{mount}} \geq f_{\text{coupling, max}} ,
    \label{eq:simpletipslip}
\end{equation}
\noindent
where $F_{v,a}$ is the axial component of the force from the vine robot body pushing away the tip mount and $W$ is the axial component of the force due to gravity acting on the tip mount. 
$f_{\text{mount}}$ is the sum of all axial frictional forces acting on the exterior of the tip mount, and $f_{\text{coupling, max}}$ is the maximum value for the friction between the tip mount and the vine robot body tail. We can also modify the growth balance model \eqref{eq:simpletetherpayload} with the new forces in Fig. \ref{fig:fbds}A to find that growth can only occur when: 
\vspace{-0.1cm}
\begin{equation}  
\begin{split}
    \frac{1}{2}PA \geq \ & F_{\text{eversion}} + T_{\text{tail}}\ + F_{\text{load}} \; + \\
& W + f_{\text{mount}} + F_{\text{interaction}}(F_{v,a}, \gamma) ,
\label{eq:simpletip}
\end{split}
\end{equation}
\noindent
where $F_{\text{interaction}}(F_{v,a}, \gamma)$ is the effective propulsion force required to produce $F_{v,a}$ and offset deformation in the vine robot tip's geometry due to interaction with the tip mount ($\gamma$). 

Critically, we expect $F_{\text{interaction}}$ to monotonically increase with $F_{v,a}$, because $F_{v,a}$ represents propulsive force wasted in pushing away the tip mount. 
An ideal constant force tip mount would therefore minimize $F_{v,a}$, which occurs when the tip mount is on the verge of slipping without contacting the vine robot tip: 
\vspace{-0.1cm}
\begin{equation}
    f_{\text{coupling,max}} = F_{\text{load}} + W - f_{\text{mount}} .
    \label{eq:mostefficienttip}
\end{equation}

In this case, only a negligible $F_{v,a}$ is needed to induce slipping \eqref{eq:simpletipslip}. Under this condition, $F_{\text{interaction}}(F_v, \gamma) \approx 0$, so the only propulsion losses between growth with a constant force enclosed tip mount \eqref{eq:simpletip} and no tip mount \eqref{eq:simplegrowth} are the added $W$ and $f_{\text{mount}}$ terms. However, this situation cannot be achieved, because the tip mount can only pull a load forward when:
\vspace{-0.1cm}
\begin{equation}
    f_{\text{coupling, max}} \geq F_{\text{load}} + W + f_{\text{mount}} .
    \label{eq:minsimpletipfriction}
\end{equation}

While the ideal value of $f_{\text{coupling, max}}$ from \eqref{eq:mostefficienttip} cannot satisfy this condition if $f_{\text{mount}} \ne 0$, a more significant issue comes from the impossibility of predicting the maximum value of $F_{\text{load}} + W + f_{\text{mount}}$. This forces $f_{\text{coupling, max}}$ to be overestimated relative to its ideal maximum in \eqref{eq:mostefficienttip} to prevent the tip mount from slipping and being left behind. However, this overestimate corresponds to an increase in propulsive force losses from $F_{\text{interaction}}(F_{v,a}, \gamma)$, reducing the maximum robot length.

\subsubsection{Passively Adapting Enclosed Tip Mount}



Unlike the constant force enclosed tip mount, the passively adaptive enclosed tip mount is designed to have a variable $f_{\text{coupling,max}}$ that decreases when the tip mount reaches the tip of the vine robot. From the free-body diagram in Fig. \ref{fig:fbds}b, by taking the moment about pivot $N$ for either arm of the tip mount, we find that $f_{\text{coupling,max}}$ decreases as $F_{va}$ and $F_{vb}$ increase and act to open the clamp:
\vspace{-0.1cm}
\begin{equation}
\begin{split}
    \lVert \overrightarrow{NS} \times \overrightarrow{F}_{s}\rVert- \lVert \overrightarrow{NA} \times \overrightarrow{F}_{va}\rVert 
= \left( \frac{d}{\mu_s}-w \right) f_{\text{coupling,max}} 
\\
\lVert \overrightarrow{NM} \times \overrightarrow{F}_{s}\rVert- \lVert \overrightarrow{NB} \times \overrightarrow{F}_{vb}\rVert 
= \left( \frac{d}{\mu_s}+w \right) f_{\text{coupling,max}} ,
\label{eq:adaptivetipmomentbalance}
\end{split}
\end{equation}
\noindent
where $F_s$ is the spring force acting to close the tip mount clamp, and $d$ and $w$ are the lengths of the moment arms for the resultant normal ($\frac{f_{\text{coupling}}}{\mu}$) and friction ($f_{\text{coupling}}$) forces acting on the tip mount clamp. The vectors $\overrightarrow{NS}$, $\overrightarrow{NA}$, $\overrightarrow{NM}$, and $\overrightarrow{NB}$ are position vectors from pivot $N$ to the point of application of the corresponding spring force ($\overrightarrow{f_s}$) and force from the vine robot body pushing away the tip mount ($\overrightarrow{F_{va}}$ and $\overrightarrow{F_{vb}}$).

Similar to \eqref{eq:simpletipslip}, the condition for the tip mount to be pushed away from the vine robot tip for growth to occur is:
\vspace{-0.1cm}
\begin{equation}
\begin{split}
    (F_{va,a} + F_{vb,a}) + F_{\text{load}} + W - f_{\text{mount}} \\
\geq f_{\text{coupling,max}}(F_{va}, F_{vb}),
\label{eq:adaptivetipslip}
\end{split}
\end{equation}
\noindent
where $F_{va,a}$ and $F_{vb,a}$ are the axial components of the reaction forces $F_{va}$ and $F_{vb}$, and the analog of $F_{v,a}$ in \eqref{eq:simpletipslip}. Unlike for the simple enclosed tip mount, $f_{\text{coupling,max}}$ is now a monotonically decreasing function of $F_{va,a}$ and $F_{vb,a}$ as seen from \eqref{eq:adaptivetipmomentbalance}. 

Also, similar to \eqref{eq:simpletip}, the growth condition for a vine robot with a passively adapting tip mount is:    
\vspace{-0.1cm}
\begin{equation}  
\begin{split}
    \frac{1}{2}PA \geq \ & F_{\text{eversion}} + T_{\text{tail}}\ + F_{\text{load}} \; + \\
& W + f_{\text{mount}} + G_{\text{interaction}}(F_{va,a}, F_{vb,a},  \gamma).
\label{eq:adaptivegrowthpressure}
\end{split}
\end{equation}
\noindent
The change in tip mount design and contact conditions yield a different function for the effective propulsion force $G_{\text{interaction}}(F_{va,a}, F_{vb,a},  \gamma)$ required to produce $(F_{va,a} + F_{vb,a})$ and account for deformation in vine robot tip geometry ($\gamma$). 

As the tip mount is pulled towards the vine robot tip, $F_{va}$ and $F_{vb}$ increase, decreasing $f_{\text{coupling, max}}(F_{va}, F_{vb})$. This in turn reduces the forces $F_{va,a}$ and $F_{vb,a}$ required to push away the tip mount and the propulsion force loss $G_{\text{interaction}}(F_{va,a}, F_{vb,a},  \gamma)$ associated with producing those forces. This effect is maximized as the sensitivity of $f_{\text{coupling,max}}(F_{va,a}, F_{vb,a})$ increases. This force reduction can be achieved by designing longer moment arms corresponding to the $F_{va,a}$ and $ F_{vb,a}$ forces, or by decreasing the moment arm $d$. As will be seen in Section \ref{section:experimentaltesting}, $f_{\text{coupling,max}}(F_{va,a}, F_{vb,a})$ can be tuned such that, for a given maximum tip mount load, the passively adapting enclosed tip mount introduces less propulsive force loss than a constant force enclosed tip mount.

\section{Experimental Testing}
\label{section:experimentaltesting}

\begin{figure*}[t]
  \centering
  \includegraphics[width=1\textwidth]{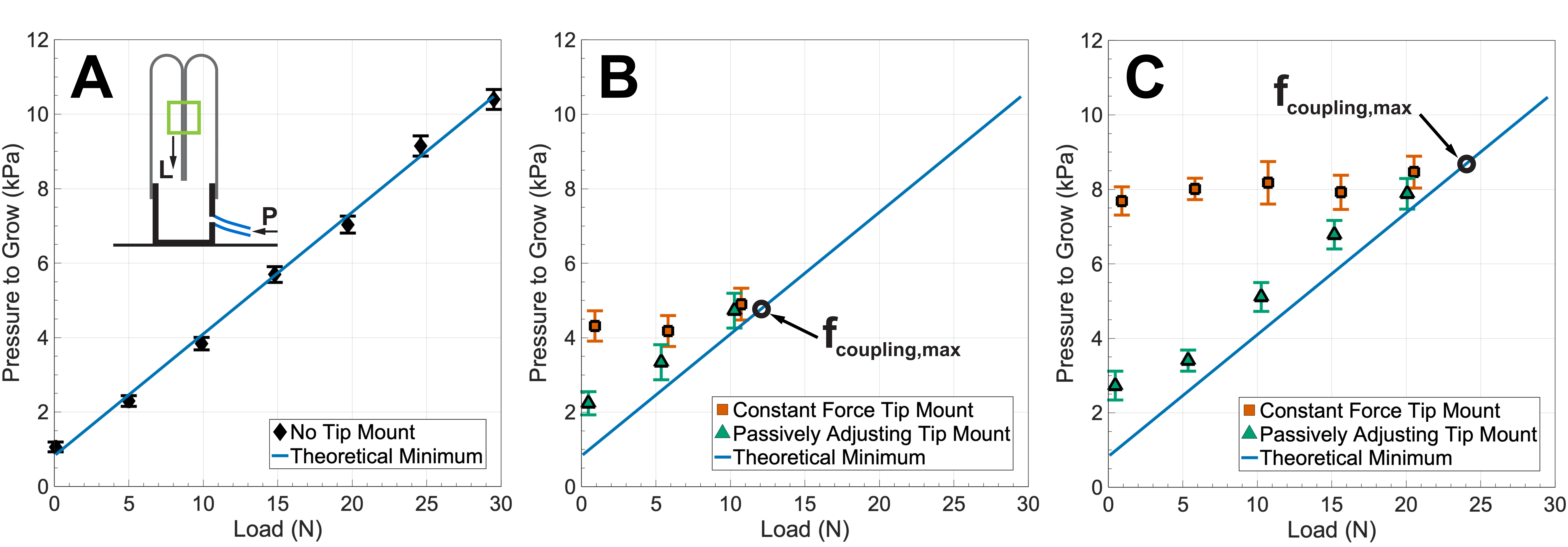}
  \caption{The pressure required for a vine robot to begin growing with various loads attached, with and without tip mounts. (A) Growth with the load directly attached to the vine robot's tail. This method requires the least pressure to grow for a given load. The experimental setup used for this and all other tests is diagrammed in the top left of the subfigure. (B) A comparison of the growth pressure required when using the constant force tip mount and passively adjusting the enclosed tip mount as attachment points. In these tests, the tip mount clamps were tuned to slip at loads above 12 N. (C) The same tests as in subfigure B, but with the tip mount maximum friction force tuned to 24 N.}
  \label{fig:experimentaldata}
\end{figure*}

To validate the models described in Section \ref{section:modeling}, we performed a series of vine robot growth tests. For all tests, we attached a series of masses to the tail of the vine robot or an attached enclosed tip mount. We then used rubber gaskets and a hose clamp to attach an 88 mm diameter vine robot body made from a 0.18 mm thick LDPE tube (Uline) to a pressure vessel. Finally, we gradually increased the pressure within the vine robot body until growth was visually observed. 

For each load, 10 trials were performed to reduce the variance from material wrinkling and variable contact conditions. The tail material was folded as described in Section \ref{section:designvinerobotbody}, and retraction was not conducted between trials to ensure unwrinkled LDPE was being everted every trial.

Pressure was measured with a differential pressure gauge (ND030D, Superior Sensor Technology, Inc.). As the vine robot was oriented to grow vertically against gravity, the load on the vine robot was the sum of the weight of the vine robot's tail, the weight of the tip mount (if applicable), and the weight of the attached masses. The axial friction forces acting on the tip mount beyond the clamping force were assumed to be zero or negligible. The vine robot tail length was approximated as 25 cm long with a corresponding mass of 9 grams.   

\subsection{Growth without Tip Mount}
\label{section:experimentalnotip}


We performed a series of tests with no tip mount to determine the absolute minimum required pressure to grow as a function of load for our specific vine robot (Fig. \ref{fig:experimentaldata}A). As this corresponds to the growth conditions in \eqref{eq:simplegrowth} and \eqref{eq:simpletetherpayload}, we were also able to experimentally determine the geometric factor $C$ in \eqref{eq:simplegrowth} and the material resistance to eversion (for our specific vine robot). We found that $C \approx \frac{1}{2}$ with an experimentally determined value of 0.503, and $F_{\text{eversion}} = 2.52 N$.

\subsection{Growth with Tip Mount}

Both a constant force enclosed tip mount and a passively adapting enclosed tip mount (Fig. \ref{fig:fbds}) were tested with varying payload masses. The clamping spring force was tuned so both tip mounts were tested with a maximum clamping force $f_{\text{coupling,max}}$ or $f_{\text{coupling, max}}(F_{v,a}, F_{v,b} = 0)$ equal to 12 N (Fig. \ref{fig:experimentaldata}B) and 24 N (Fig. \ref{fig:experimentaldata}C). The tip mounts' maximum coupling friction with the vine robot tail was measured with a force gauge between sets of trials and kept within 10\% of the nominal value. 

In both graphs, we see that as the load approaches the maximum coupling friction, the pressure required to grow approaches the theoretical optimum. This agrees with the prediction from \eqref{eq:mostefficienttip}, since less interaction force $F_{v,a}$ or $(F_{va,a} + F_{vb,a})$ is required to push the tip mounts away from the vine robot tip. However, as the load decreases, we see an increasing divergence between the constant force and passively adjusting enclosed tip mounts. 

For the constant force enclosed tip mount, as the load force $F_{\text{load}}$ decreases, the required interaction force increases as modeled in \eqref{eq:simpletipslip}. Thus, the decrease in $F_{\text{load}}$ in the growth condition equation \eqref{eq:simpletip} is offset by the need for a larger $F_{\text{interaction}}(F_{v,a}, \gamma)$. For the constant force tip mount design tested, this offset resulted in nearly constant pressure to grow.

For the passively adapting tip mount, though a decreasing $F_{load}$ leads to an increase in required force to push away the tip mount ($F_{va,a} + F_{vb,a})$ \eqref{eq:adaptivetipslip}, this is partially offset by the larger interaction forces $F_{va}$ and $F_{vb}$ decreasing $f_{\text{coupling, max}}(F_{va}, F_{vb})$ \eqref{eq:adaptivetipmomentbalance}. Instead of a nearly constant pressure to grow, we see an offset from the theoretical minimum pressure that only gradually increases as $F_{\text{load}}$ decreases. 

While both the constant force and passively adaptive tip mount approach the theoretical minimum pressure as load approaches the maximum coupling force, this is not an ideal operating regime due to the risk of the tip mount no longer being pulled forward. If instead a safety factor for the maximum coupling friction is used, the passively adaptive enclosed tip mount introduces less propulsion force losses, allowing the vine robot to grow further by overcoming larger values of $T_{\text{tail}}$ for the same maximum internal pressure. 

\section{Demonstrations}

\subsection{Laboratory Testing}

\begin{figure*}[t]
  \centering
\includegraphics[width=1.00\textwidth]{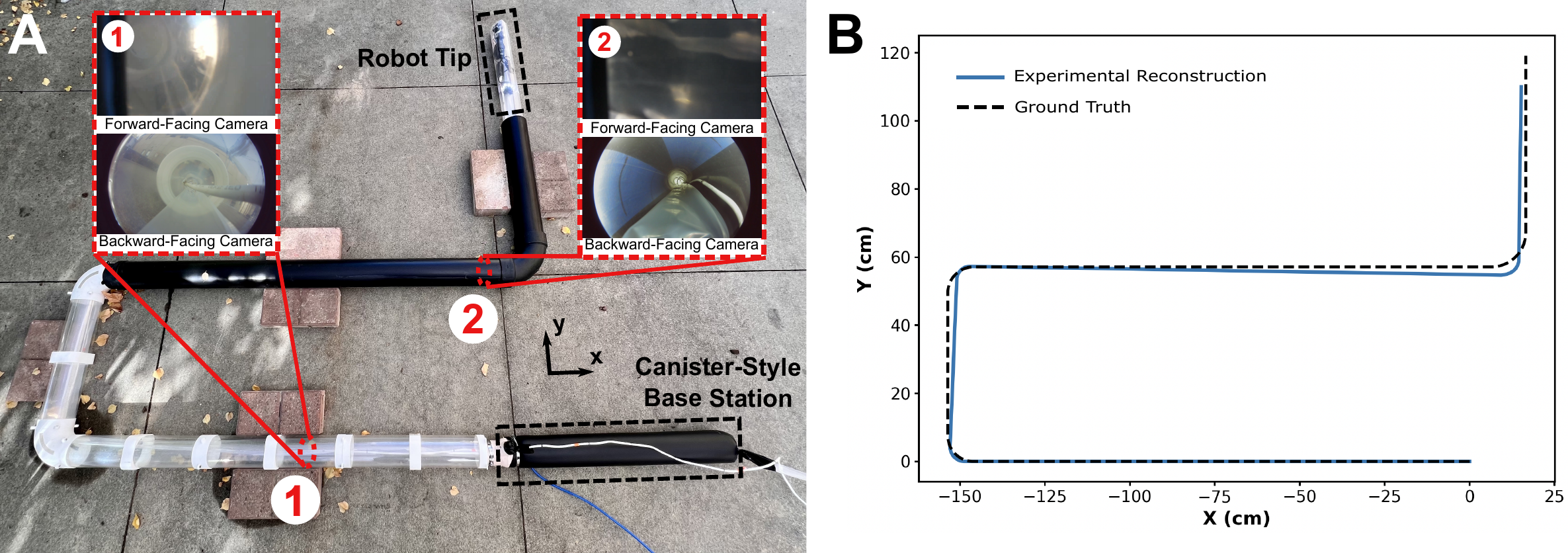}
  \caption{Laboratory testing of the hermetic and transparent vine robot system was performed in a 10 cm inner diameter, 4.57 m long pipe system. (A) Growth through a pipe system, with inset images (1) and (2) showing the view from the backwards and forwards-facing cameras at corresponding locations in the pipe. (B) A visualization of the robot's path compared to the ground truth path.}
  \label{fig:labtest}
\end{figure*}

To perform initial system validation and testing, we constructed a 4.57 m long pipe with 3x 90 degree bends from 10 mm ID PVC and PETG tubes. We grew our enclosed vine robot system through the pipe while recording localization and visual data (Fig. \ref{fig:labtest}). The vine robot system weighed 600 g not including the Ethernet cable, and initially grew at a pressure of approximately 3.4 kPa. 

We found that the backward-facing camera consistently provided high quality images, with partial occlusion from the vine robot tail and tether. The forward-facing camera was often fully obscured by everting vine robot material or partially obscured by an arm of the tip mount. In Fig. \ref{fig:fieldtest}, image set (1) shows a forward-facing camera image that is relatively unobscured by the vine robot body, but partially blocked by the tip mount. Image set (2) shows a forward-facing camera image that is obscured by everting vine robot material. However, we did not deem this a critical issue as the backwards camera could detect large changes in pipe features.

From the path reconstruction results shown in Fig.~\ref{fig:labtest}B, the 2D experimental trajectory closely matches the ground-truth pipe configuration. The reconstructed path exhibits less than 2.6 degrees of orientation deviation and less than 10 cm of overall length deviation. The observed orientation errors are likely due to inherent sensor drift, particularly under magnetic disturbance, and undesired tilting of the sensor mount. The length deviation primarily arises from odometry inaccuracies, as the Ethernet cable does not perfectly follow the pipe’s centerline, particularly through elbow sections, and from potential errors in manual marker labeling.

\subsection{Field Deployment}


To test our hermetic and transparent vine robot system in a real-world scenario, we collaborated with a chemical company on a pipe inspection and mapping challenge. The task involved a 10 cm (4 in) diameter wastewater pipe which transported sewage away from a lift station. A section of the pipe had failed, and the company wanted to determine whether trenchless repair options could be used. However, because of outdated maps and the potential for cracks, wall collapse, or sunken pipe sections, basic information about the path and internal condition of the pipe was needed. The company specifically desired visual data of the inside of the pipe with enough quality to detect large defect or failures in the pipe walls, and a map of the pipe accurate to within 1 meter in case trenching was required. While the pipe would be flushed out, there could be sunken sections filled with water or biohazards. Additionally, the placement of the pipe and pump station meant that using the pump station as an access point would be highly disruptive. 

\begin{figure*}
\begin{center}
\includegraphics[width=0.9\textwidth]{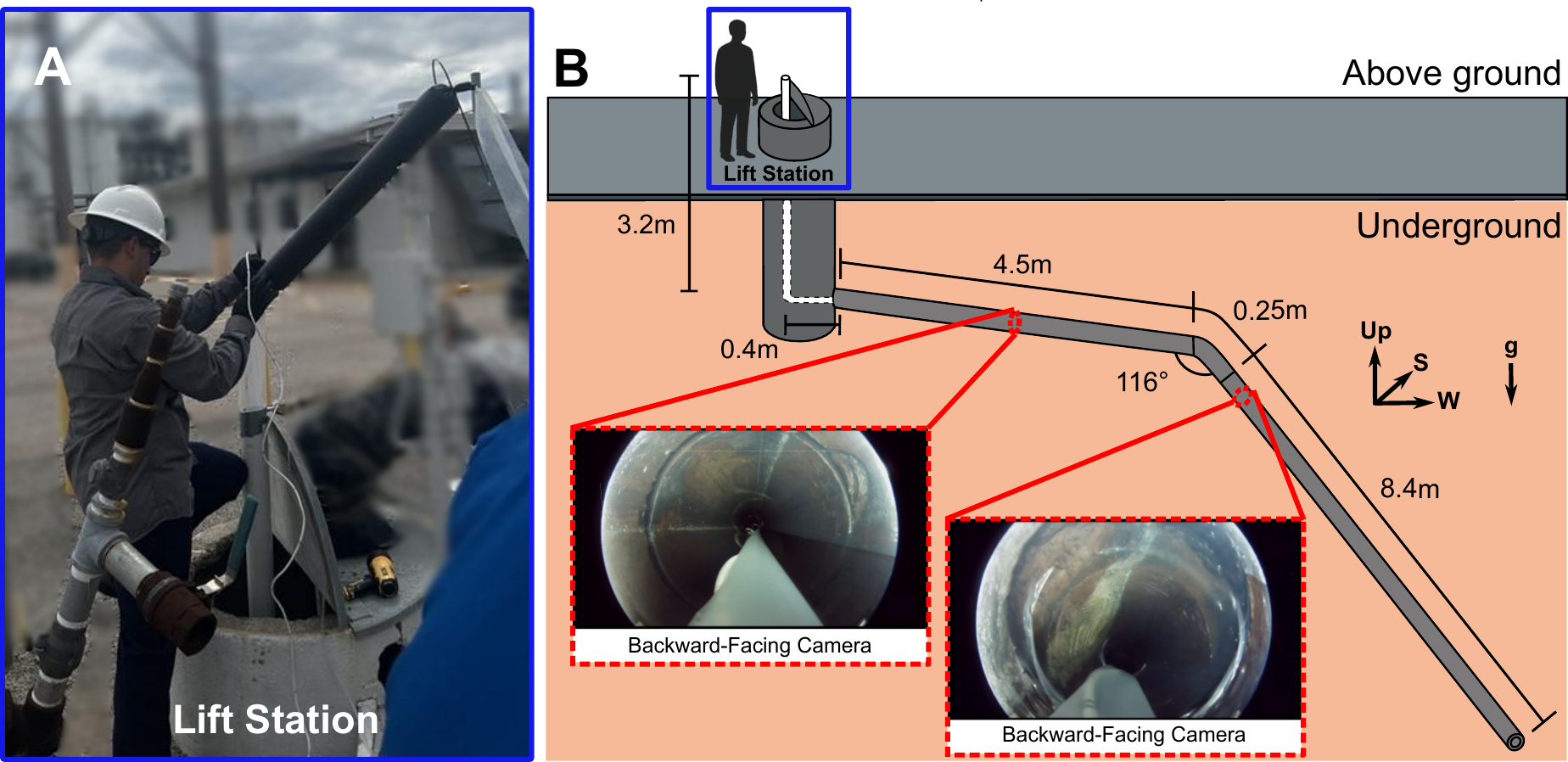}
  \caption{Results from a field test conducted in a wastewater pipe system. (A) The system being deployed into a sewage lift station through a vertical PVC pipe spool. (B) A visualization of the mapping data collected with the robot's sensors and manual odometry. The inset images show the view from the robot’s rear-facing camera as it grew through the wastewater pipe.}
  \label{fig:fieldtest}
\end{center}
\end{figure*}

The company decided that the pipe's small diameter and unknown internal environment made commercial inspection robots too risky and costly to try, while borescopes also risked being limited if multiple turns were present. Our system provided the potential for a lower risk inspection, so we were give the opportunity to deploy in the wastewater pipe.
The results of our field deployment are shown in Fig. \ref{fig:fieldtest}. The robot was primarily operated by an employee of our collaborator company, and proved easy to operate with minimal training or instruction. The robot traversed a 3.6 m long, 7.6 cm (3 in) inner diameter pipe spool which was inserted into the entrance of the wastewater pipe. Based on localization data, the robot then traversed a 4.5 m long section of pipe running East to West at an 8.7 degree angle of depression, a 0.25 m long 116 degree elbow, and an 8.4 m long section of pipe at a 6.2 degree angle of depression (Fig. \ref{fig:fieldtest}B). The deployment concluded when the company believed the robot had collected enough information to de-risk future rehabilitation operations. The inset images in Fig. \ref{fig:fieldtest}B provide two examples of images from the backwards facing camera, one showing damage in the pipe wall, and one showing the robot rounding a turn.

\section{Conclusions}
In this work, we presented a hermetic and transparent vine robot system for vision-based inspection of junction-less pipes. By designing for a common but still challenging set of pipe inspection tasks, we realized a robust, streamlined, field-deployable system. Our vine robot uses a transparent LDPE vine robot body, one actuator (an air compressor), one mechanism (a hinge), and minimal electronics. It requires no specialized measures to prevent water damage, fouling, or interference with the environment. Because the hermetic and transparent system introduced the need for a fully enclosed tip mount to keep sensors at the tip of the vine robot, we developed an unactuated, passively adapting design that  significantly reduces propulsion losses and enables further travel within pipes. In both laboratory and field tests, we validated that the system was robust, easy to operate, and could collect a basic level of condition sensing and mapping data. 

In future work, we will improve backward- and forward-facing camera positions by having more of each and positioning them around the vine robot tail and farther from the tip of the vine robot, respectively. Additionally, we will replace the error-prone dead-reckoning used here with an exteroceptive sensing method such as optical flow or structured light to reduce drift and improve localization accuracy. Finally, we note that many existing technologies such as internal vine robot steering mechanisms \cite{internal_reeling_steering}, non-contact sensors such as eddy current and magnetic flux leakage, and distributed sensor networks \cite{dist_sensor} can be incorporated into a hermetic and transparent vine robot system to increase its capabilities while retaining the advantages of a fully enclosed robotic system.



\bibliographystyle{ieeetr}
\bibliography{main}

\end{document}